\documentclass[letterpaper]{article} 
\usepackage{aaai2026}  
\usepackage{times}  
\usepackage{helvet}  
\usepackage{courier}  
\usepackage[hyphens]{url}  
\usepackage{graphicx} 
\usepackage{natbib}  
\usepackage{caption} 
\nocopyright
\usepackage{newfloat}
\usepackage{listings}
\DeclareCaptionStyle{ruled}{labelfont=normalfont,labelsep=colon,strut=off} 
\title{Explicit World Models for Reliable Human-Robot Collaboration}
\author {
   Kenneth Kwok\textsuperscript{\rm 1},
    Basura Fernando\textsuperscript{\rm 1},
    Qianli Xu\textsuperscript{\rm 2},\\
   Vigneshwaran Subbaraju \textsuperscript{\rm 1},
    Dongkyu Choi\textsuperscript{\rm 3},
    Boon Kiat Quek\textsuperscript{\rm 1}
   }
\affiliations {
    \textsuperscript{\rm 1}Institute of High Performance Computing, Agency for Science, Technology and Research, Singapore\\
    \textsuperscript{\rm 2}Institute for Infocomm Research (I\textsuperscript{2}R), Agency for Science, Technology and Research, Singapore\\
        \textsuperscript{\rm 3}A-FAB Technology Team, Mechatronics Research, Samsung Electronics, Korea\\
    kenkwok@a-star.edu.sg,fernando\_basura@a-star.edu.sg, xu\_qianli@a-star.edu.sg,\\
    vigneshwaran\_subbaraju@a-star.edu.sg, edc.choi@samsung.com, quekbk@a-star.edu.sg
    
}

\begin{document}

\maketitle


\section*{Overview}
This paper addresses the topic “\textit{Robustness under sensing noise, ambiguous instructions, and human-robot interaction}”. We take a radically different tack to the issue of reliable embodied AI: instead of focusing on formal verification methods aimed at achieving model predictability and robustness, we emphasise the dynamic, ambiguous and subjective nature of human-robot interactions that requires embodied AI systems to perceive, interpret, and respond to human intentions in a manner that is \textbf{consistent, comprehensible and aligned with human expectations}. We argue that when embodied agents operate in human environments that are inherently social, multimodal, and fluid, reliability is contextually determined and only has meaning in relation to the goals and expectations of humans involved in the interaction. This calls for a fundamentally different approach to achieving reliable embodied AI that is centred on building and updating an accessible \textit{explicit world model} representing the common ground between human and AI, that is used to align robot behaviours with human expectations. 
\section*{Building Common Ground}
\paragraph{Human Inspiration}
Humans learn to interpret the world not only through static visual or speech perception, but through \textit{continuous} \textit{integration} of multimodal cues including gaze, gestures, prosody and movement dynamics, and contextual knowledge. These cues carry rich inferential signals about what others mean, what they want, and what will happen next, and contribute to developing a mutual understanding of the world that forms the basis for cooperative action. This shared conception of the world has been termed \textit{common ground}, a joint understanding of the tasks, communications, and environments between agents \cite{Dillenbourg1999GroundingIM}.
\paragraph{Common Ground}
The idea of common ground originates from language and cognition studies \cite{Clark1983CommonGA} and has been extensively studied in the field of human-AI teaming under the hood of shared mental models, which cover constructs such as knowledge representation, schema, and situation awareness \cite{Andrews2022TheRO}. In the domain of Human-Robot Collaboration (HRC), this concept underpins effective teamwork, requiring sophisticated mechanisms to bridge differences between human and artificial agents in terms of perception, cognition, and embodiment \cite{Tan2020TaskOrientedMQ}.

\paragraph{Perceptual Grounding}
Perceptual grounding is arguably the first step in the establishment of common grounds to construct a valid world model for HRC. Research in this domain has been centred around visual understanding enabled by deep learning models, and more recently visual foundational models. These models have been used to address various tasks and benchmarks on Visual Question Answering (VQA) (Zhong et al. 2022), which nevertheless is inadequate at capturing the dynamic, multimodal, and task-specific context of HRC. Therefore, a growing interest is observed in building common grounds for \textit{task-oriented} collaborations.  We proposed a Task-oriented Collaborative Question Answering (TCQA) benchmark \cite{Tan2020TaskOrientedMQ} for benchmarking grounding methods with quantitative evaluation of their effectiveness in HRC tasks. Our baseline model combining deep learning to tackle basic perception and symbolic reasoning to capture high-level contextual information and reasoning achieved good performance on the benchmark, but this approach still suffers from fragility/errors in novel scenes and lacks flexibility in constructing new semantic inferences. To address these issues, Large Language and Multimodal Models (LLMs/LMMs) have been leveraged for semantic knowledge to inform affordance reasoning \cite{Ahn2022DoAI,Huang2023VoxPoserC3}, coordination \cite{Zhang2024TowardsEL} and human goal reasoning \cite{Wan2023HandMeThatHC}.  However, these approaches face challenges owing to their intrinsic disembodiment from the physical world.

\paragraph{Joint Attention and Multimodal Interaction}
Foundational work in social robotics has emphasised the importance of joint attention and shared intentionality for meaningful interaction. Scassellati demonstrated that \textit{joint attention} enables robots to interpret human referential cues \cite{scassellati1996mechanisms}. Extending this, Breazeal et al. showed that \textit{non-verbal behaviours} significantly improve efficiency and robustness in human–robot teamwork \cite{breazeal2005effects}, revealing that embodied communication is essential for reliable coordination, while Sato et al. showed that continuous monitoring of human behaviours expressed both via conscious actions/language and unconscious/involuntary non-verbal cues is needed for robots to actively \textit{infer human intentions} \cite{sato1995active}.  In parallel, work on legible robot motion showed that robots must act not just efficiently, but also \textit{expressively}, producing behaviours that communicate intent to human partners, improving predictability and coordination in shared workspaces \cite{dragan2013legibility}.  These works collectively support the argument that reliable collaboration emerges from interactive common ground building, not merely isolated perception.
In related work, we demonstrated how multimodal human cues are essential for reliable referential grounding. In M2GESTIC \cite{weerakoon2020gesture}, we showed that a distance-weighted understanding of pointing gestures can significantly reduce ambiguity in comprehending natural multi-modal human instructions. We also demonstrated that eye gaze provides strong cues for predicting referents and action steps during joint tasks \cite{johari2021gaze}. COSM2IC \cite{weerakoon2022cosm2ic} introduced adaptive real-time multimodal fusion that prioritises gesture or linguistic structure depending on context, highlighting that reliability emerges from dynamic coordination rather than rigid pipelines. Most recently, Ges3ViG \cite{mane2025ges3vig} integrates pointing gestures with 3D visual grounding, advancing spatially grounded reference understanding for real-world embodied AI. 

\section*{Explicit World Models}
\paragraph{Cognitive Architectures (CAs)}
These symbolic AI systems rely on symbol manipulation and reasoning based on logical rules emulating human cognitive processes and have traditionally used explicit models of the world to represent the environment, relational concepts, and executable procedures. Symbols and their assigned semantics are formally defined within the coherent structure of world models that are accessible by the agent's cognitive processes.
The central challenge in constructing an explicit world model is the representation of environmental state descriptions, which requires formalisms capable of representing the facts about states. Logical languages like first-order predicate logic are often used to directly represent entries within a knowledge representation formalism. Such representations enable symbolic systems to transform raw sensory data into a high-level, interpreted, and structured abstraction of the environment, making the information accessible for cognitive processing. However, most symbolic approaches are heavily dependent on human handcrafting and are unable to scale to represent the complexities of real worlds.

\paragraph{Neuro-Symbolic Architectures}
More recently, we observe the emergence of explicit, interpretable, and self-evolving world models as the foundation for next-generation neuro-symbolic intelligence. Weng’s 
early insights into autonomous mental development \cite{Weng2012AutonomousMentalDev} framed a crucial distinction between symbolic and emergent representations, arguing that genuine intelligence requires the brain—or its artificial counterpart—to develop its internal representations without human handcrafting, forming abstractions directly from sensorimotor experience.  In essence, Weng’s vision anticipated the need for agents that construct and continually refine internal world models capable of linking sensory input to motor behaviour and abstract reasoning—a theme that now defines modern neuro-symbolic research.  
\paragraph{}
Consider for example NeSyC \cite{Choi2025NeSyC}, a neuro-symbolic continual learner inspired by the hypothetico-deductive model of scientific reasoning. It combines the generative creativity of large language models (LLMs) with the logical precision of symbolic solvers, creating a feedback loop where inductive inference (via LLMs) and deductive validation (via Answer Set Programming) reinforce each other. Through contrastive learning and continual memory refinement, NeSyC can generalise actionable knowledge across diverse open-domain environments — transforming raw experience into structured, symbolic understanding. Our recent work \cite{Nguyen2025KML} on Knowledge Module Learning (KML) and the PKR-QA benchmark for procedural reasoning deepens this trajectory. We encode procedural knowledge in a knowledge graph linking tasks, steps, actions, objects, tools, and purposes, grounding symbolic reasoning in perceptual and temporal context. KML trains neural knowledge modules to capture relations between entities — bridging the gap between statistical learning and symbolic compositionality. When combined with LLM-generated reasoning programs, these modules yield interpretable, stepwise reasoning traces that can be verified and debugged. The integration of structured knowledge with neural embeddings transforms procedural understanding from mere sequence prediction into causal reasoning — enabling agents to explain why a particular action should occur, not just what should be done next.

\section*{Conclusion and a Call to Action}
\paragraph{}
Current trends in embodied AI favour end-to-end approaches to learn black-box models for controlling robots.  Such approaches place the burden of reliability entirely upon learning verifiably correct models for different tasks and situations.  For HRC, this might not be viable given the inherently dynamic, ambiguous and subjective nature of human-robot interactions.  We argue instead that reliable collaborative behaviour needs to be constructed \textit{on-the-fly} through mutual building and maintenance of an explicit world model to serve as common ground between humans and robots.
\paragraph{}
World models resolve ambiguity and subjectivity through explicit commitment to interpretations of environmental states and human intentions.  But for this to work, they need to be light-weight enough for real-time updating,
yet sufficiently representative to capture the rich social, multimodal, and fluid nature of interactions between humans and robots.

\paragraph{}
In this position paper, we reviewed related work in human-inspired construction of common ground from rich human-robot interactions, and in explicit world modelling in AI systems, to motivate a shift from opaque models to explicit world models that can provide common ground for guiding reliable collaborative behaviours in human-robot teams.  Challenges to realising this approach that have been identified will require  multidisciplinary contributions from the diverse communities present in this Bridge to solve.

\bibliography{aaai2026}


\end{document}